\documentclass[10pt,twocolumn,letterpaper]{article}

\usepackage[pagenumbers]{cvpr} % To force page numbers, e.g. for an arXiv version

\usepackage{graphicx}
\usepackage{amsmath}
\usepackage{amssymb}
\usepackage{booktabs}
\usepackage{multirow}
\usepackage{enumitem}

\usepackage[table,xcdraw]{xcolor}
\usepackage{mathtools}
\usepackage{adjustbox}
\usepackage{indentfirst}
\usepackage{multirow}
\usepackage[T1]{fontenc}
\usepackage{mathrsfs}
\usepackage{array}
\usepackage{colortbl}
\usepackage{bm}
\usepackage{lipsum}
\usepackage{color}

\usepackage[pagebackref,breaklinks,colorlinks]{hyperref}

\usepackage[capitalize]{cleveref}
\crefname{section}{Sec.}{Secs.}
\Crefname{section}{Section}{Sections}
\Crefname{table}{Table}{Tables}
\crefname{table}{Tab.}{Tabs.}

\def\cvprPaperID{5086} % *** Enter the CVPR Paper ID here
\def\confName{CVPR}
\def\confYear{2022}

\begin{document}

\title{Semi-Supervised Wide-Angle Portraits Correction by Multi-Scale Transformer}
\author{Fushun Zhu\,$^1$\thanks{Equal contribution. $^{\dagger}$Corresponding authors.}\quad
Shan Zhao\,$^{2*}$\quad
Peng Wang\,$^{2*}$\quad
Hao Wang\,$^2$\quad
Hua Yan\,$^{1\dagger}$\quad
Shuaicheng Liu\,$^{3,2\dagger}$  \\
$^1$\,Sichuan University\quad$^2$\,Megvii Technology\\
$^3$\,University of Electronic Science and Technology of China\\
}
\maketitle

% \begin{document}
% %%%%%%%%% TITLE
% \title{Semi-Supervised Wide-Angle Portraits Correction by Multi-Scale Transformer}

% \author{
%   Fushun Zhu\affmark[1]\thanks {Equal contribution} \quad 
%     Shan Zhao\affmark[2]\footnotemark[1] \quad 
%     Peng Wang \affmark[2]\footnotemark[1] \quad  
%     Hao Wang \affmark[2]\footnotemark[1]  \quad 
%     Hua Yan   \affmark[1]\footnotemark[1]  \quad 
%     Shuaicheng Liu\affmark[2,3] \thanks {Corresponding author}\\\\

% \affaddr{\affmark[1]Sichuan University} \quad
% \affaddr{\affmark[1]Megvii Research} \quad
% \affaddr{\affmark[2]University of Electronic Science and Technology of China}\\

% {  
%   \tt\small  zhufushun@stu.scu.edu.cn, yanhua@scu.edu.cn, liushuaicheng@uestc.edu.cn} \\
%   \tt\small \{zhaoshan, wangpeng04, wanghao03\}@megvii.com\\
% \tt\small \url{https://github.com/megvii-research/Portraits_Correction}
% }

%%%%%%%%% ABSTRACT
\begin{abstract}
We propose a semi-supervised network for wide-angle portraits correction. Wide-angle images often suffer from skew and distortion affected by perspective distortion, especially noticeable at the face regions. Previous deep learning based approaches need the ground-truth correction flow maps for training guidance. However, such labels are expensive, which can only be obtained manually. In this work, we design a semi-supervised scheme and build a high-quality unlabeled dataset with rich scenarios, allowing us to simultaneously use labeled and unlabeled data to improve performance. Specifically, our semi-supervised scheme takes advantage of the consistency mechanism, with several novel components such as direction and range consistency (DRC) and regression consistency (RC). Furthermore,  different from the existing methods, we propose the Multi-Scale Swin-Unet (MS-Unet) based on the multi-scale swin transformer block (MSTB), which can simultaneously learn short-distance and long-distance information to avoid artifacts.  Extensive experiments demonstrate that the proposed method is superior to the state-of-the-art methods and other representative baselines.
 The source code and dataset are available at  \tt\small \url{https://github.com/megvii-research/Portraits_Correction}
 \end{abstract}

%%%%%%%%% BODY TEXT
\section{Introduction}
\label{sec:intro}
\begin{figure}[t]
	\begin{center}
		\includegraphics[width=0.95\linewidth]{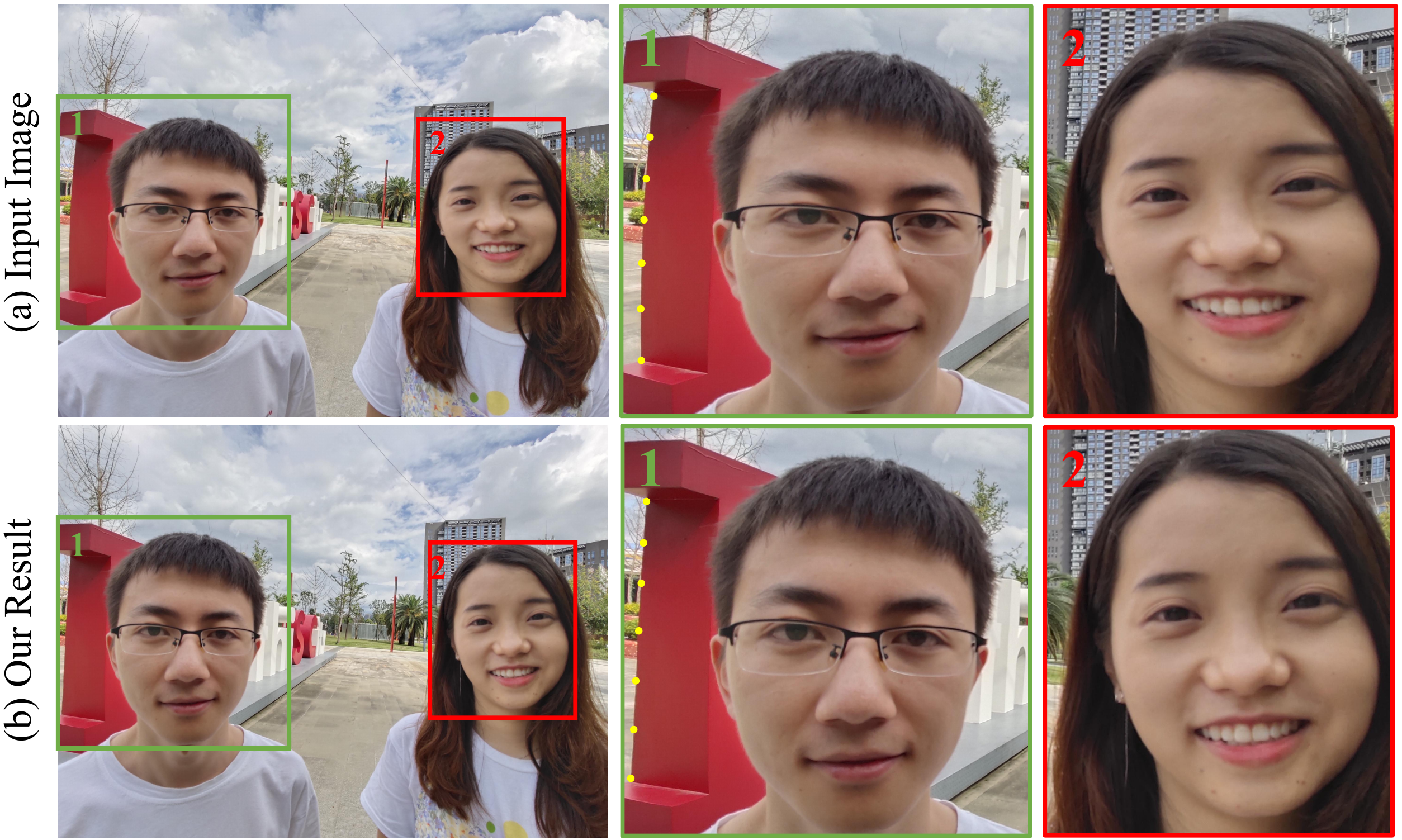}
	\end{center}
    \setlength{\abovecaptionskip}{-0.1cm}
	\caption{An example of our method. (a) the original wide-angle image with curved lines and distorted faces. (b) result by the proposed semi-supervised method, both lines and faces are corrected. }
	\label{fig:example}
\end{figure}

In recent years, a growing number of smartphones have been equipped with wide-angle cameras, which take wide-angle images with rich contents. However, a wider FOV camera often causes severe perspective distortions, which bends straight edges on buildings, and distorts faces, as shown in Fig.~\ref{fig:example}(a). Therefore, an ideal intelligent algorithm is required to correct the distortion image. After correction, the faces will look more natural while the curved lines in the background are also corrected, as shown in Fig.~\ref{fig:example}(b).
%To solve these problems, some classic calibration-based methods~\cite{pavic2006interactive,carroll2010artistic,du2013changing,tehrani2016correcting} were proposed and the curved straight lines at the background can be corrected. But these conventional methods fail to correct the distorted faces, and even have a negative impact. Recent approaches~\cite{shih2019distortion,tan2021practical} have considered to correct both distorted portraits and lines simultaneously. Especially, the first supervised CNN-based method was proposed by Tan \emph{et al.}~\cite{tan2021practical} and obtained satisfactory result. In Tan's method, two networks were designed to remove the distortion on background and portraits respectively, achieving smooth transitions between perspective-rectified background and stereographic-rectified faces. Since there is no suitable dataset for training, they built a high-quality dataset over $5,000$ labeled training samples. Nevertheless, there are still some drawbacks that restrict the continued ascension in their work. First, a well-generalized network needs a large amount of training data with rich types. The high cost makes it unrealistic to enlarge the labeled dataset.
%Second, in their two-stage network, the lines at the background are required to be corrected before processing the faces, causing the existence of redundancy. Third, although CNNs have achieved excellent performance, they always rely on the receptive fields and can't learn long-range semantic information interaction well due to the locality of convolution operation.

The traditional undistortion methods apply perspective projection using calibrated camera parameters, which correctly warp the lines at the background to straight ~\cite{pavic2006interactive,carroll2010artistic,du2013changing}. Nevertheless, faces on the image are stretched unnaturally due to incorrect projection as a plane. Compared to perspective projection, the mercator and stereographic projections~\cite{svardal2003stereographic} can preserve the shape of faces locally, but they also bend linear structures in the background~\cite{carroll2009optimizing}. 

It is obvious that facial regions and background need two different types of projections for the wide-angle image correction. Carroll \emph{et al.}~\cite{carroll2009optimizing} presented a content-preserving approach that finds an optimal mapping solution according to the user-specified lines. Recently, Shih \emph{et al.}~\cite{shih2019distortion} designed an optimization problem to create a mesh that adapts stereographic projection on facial regions regionally and applies perspective projection on background, enabling a smooth transition between portraits and background by solving the optimization problem automatically. However, the method~\cite{shih2019distortion} sometimes causes distorted architectures nearby corrected faces. In addition, it requires portraits segmentation mask and camera parameters as additional inputs.

Tan \emph{et al.}~\cite{tan2021practical} proposed the first fully-supervised CNN-based method for wide-angle image correction, which consists of a line correction network and a portraits correction network. Tan's method obtained satisfactory results with the distorted image as input. However, there still exists disadvantages in their work. First, it needs many training photos under rich scenarios, and each face in the photo must be manually undistorted by specific tools. Meanwhile, errors may occur in manual annotation, causing uneven annotation quality or introducing dirty data. Therefore, the whole data preparation procedure is complex and expensive, making it unrealistic to improve performance by enlarging the training dataset. Second, Tan's method also creates artifacts in some cases because it does not use long-range semantic information for local variations of faces.
To address the above problems, we attempt to leverage a novel semi-supervised strategy, aiming to reduce the cost of preparing an expensive manual corrected dataset. Specifically, we adopt the semi-supervised strategy, containing direction and range consistency (DRC) and regression consistency (RC), to make full use of both labeled and unlabeled data by introducing a surrogate task (segmentation). Besides, compared with Tan \emph{et al.}~\cite{tan2021practical}, we develop a novel network based on the multi-scale swin transformer block (MSTB), dubbed as Multi-Scale Swin-Unet (MS-Unet) which is better suitable for portraits correction. In particular, we also collect more than $5,000$ unlabeled distortion images from different phones and scenes to train MS-Unet by the semi-supervised strategy. Experimental results show that our approach can correct distortions in wide-angle portraits with superior performance than previous methods, and it only needs a small amount of manually labeled data. 
% %通过进行了充分的实验，我们证明了所提出方法的优越性能。总结一下，我们主要的贡献包括：1.第一个用于解决畸变矫正的半监督学习方法，降低了训练数据获取的成本和难度。2.一种新颖的适用于畸变矫正任务的transformer结构。3.提供一个可用于畸变矫正的半监督学习的高质量数据集。
In summary, our main contributions are:

\begin{itemize}
\item We propose the first semi-supervised learning strategy for wide-angle portraits correction, which dramatically reduces the requirement of labeled training data.
\item We develop a novel transformer-based network called MS-Unet, based on MSTB, to fully utilize both local-scale and long-range semantic information interaction for wide-angle portraits correction.
\item We provide a high-quality unlabeled dataset that can be used to train semi-supervised wide-angle portraits correction algorithms.
\end{itemize}

\section{Related Works}
\subsection{Wide-Angle Portraits Correction}
%早期的人像畸变矫正方法主要是传统的算法。举个例子，球极投。通过球极投影可以得到非常自然的人像，但是背景上的线条会产生弯曲。更进一步的，Shih等人提出的基于网格的人像畸变方法，可以在背景直线与人像中间取得平衡，但是，它需要相机参数和一个不错的人像分割结果。最近，谭婧等人在近日提出的通过深度神经网络来实现人像畸变矫正的方法，解决了需要相机参数和依赖人像分割结果的问题。但是，该方法需要人为筛选和处理标签，导致制作标签成本大。因此，我们提出了一个半监督的transformer方法，可以低成本扩充大量训练数据，并有效提高人像畸变矫正的效果。

%人像畸变矫正技术在广角镜头的推广和应用中发挥着不可替代的作用。Portrait distortion correction plays an irreplaceable role in the popularization and application of wide-angle lenses. 
Early wide-angle portraits correction methods always relied on traditional algorithms~\cite{zorin1995correction,carroll2009optimizing}. Tehrani \emph{et al.}~\cite{tehrani2013undistorting,tehrani2016correcting} presented methods to remove faces distortions and preserve background features during this process, but their solutions require user assistance. Shih \emph{et al.}~\cite{shih2019distortion} proposed a mesh-based algorithm that can strike a balance between straight lines and faces correction effects automatically. Nevertheless, it requires the camera parameters and portraits segmentation as inputs. Recently, Tan \emph{et al.}~\cite{tan2021practical} proposed a two-stage deep neural network to complete wide-angle portraits correction with only an image as input. 
However, this fully-supervised method is limited to the number of labeled data that requires high-cost manual screening and processing. Fortunately, our method greatly reduces the limitation of the amount of labeled training dataset and learns the correction flow maps from distortion image to usual image.

\subsection{Deep Semi-Supervised Learning}
%半监督学习属于无监督学习（没有任何标记的训练数据）和监督学习（完全标记的训练数据）之间的状态，使用有标签数据+无标签数据混合成的训练数据训练网络。半监督的学习方式被广泛的应用在图像处理中，包括图像分割，图像分类，目标检测，人群计数等等。过往的这些研究也表明，这种数据混合使用的方式可以显着提高学习的准确性。因此，我们也提出了一种半监督的学习方式，打破人像畸变矫正任务中受到的训练数据量的限制。
Deep semi-supervised learning provides a practical and effective approach to fully utilizing the mixture dataset containing labeled and unlabeled images. It has been widely used in image classification~\cite{yalniz2019billion,xie2020self,he2019bag}, semantic segmentation~\cite{xiao2018transferable,babakhin2019semi,zhang2018context}, machine translation~\cite{he2016dual,cheng2019semi,fan2021beyond}, crowd counting~\cite{liu2020semi,meng2021spatial}, text classification~\cite{karamanolakis2019leveraging,li2019learning,lee2021salnet}, text segmentation~\cite{textseg, tang2017scene} and so on. These works have proved that the semi-supervised method can promote the accuracy. Therefore, we introduce the semi-supervised strategy into the portraits correction domain and make a beautiful breakthrough.

\begin{figure*}[t]
	\begin{center}
	 \includegraphics[ width=0.94\textwidth]{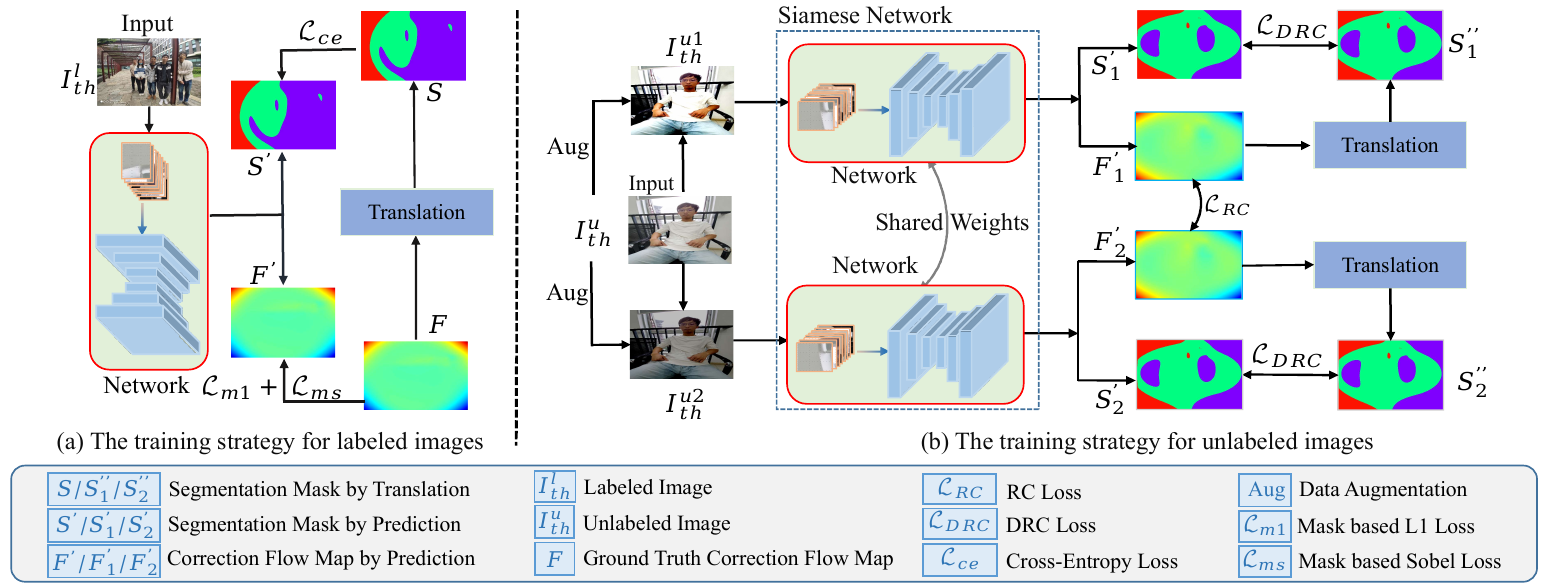}
	\end{center}
	\setlength{\abovecaptionskip}{-0.1cm}
	\caption{The pipeline of semi-supervised wide-angle portraits correction framework with the surrogate task (segmentation). (a) The network training strategy by utilizing the labeled images. (b) Utilize the unlabeled images to train our network. The training strategy consists of direction and range consistency (DRC), regression consistency (RC). For an unlabeled image $I_{th}^{u}$, when it is sent to the siamese network, the estimated segmentation mask and the correction flow map are utilized to compute the DRC loss $\mathcal{L}_{DRC}$ and RC loss $\mathcal{L}_{RC}$.}
	\label{fig:submodels}
\end{figure*}

\subsection{Visual Transformer}
%（删除了的内容）图像分类的模型ViT，它在多个图像识别任务上取得了优异的性能。与基于 CNN的方法相比，ViT的缺点是它的计算复杂度很高。为了减轻训练 ViT 的难度，出现了一些基于 ViT 改进的优秀作品（...）。其中,the image classification model ViT~\cite{dosovitskiy2020image}, which has achieved good performance on multiple image recognition tasks. Compared with CNN-based methods, the disadvantage of ViT is its high computational complexity. To reduce the difficulty of training ViT, some works based on ViT have appeared~\cite{}. Among them, 
%实现了类似CNN中的下采样的效果， achieving an effect similar to downsampling in CNN,

%Transformer的提出，深刻地改变了NLP领域。受到Transformer模型在NLP领域上取得的重大成果的启发，近几年研究人员逐渐将Transformer应用于计算机视觉领域。特别是，liu等提出了一种被称为Swin Transformer的高效的Transformer结构。它在Transformer层之间采用一种patch merge的机制将相邻的patch特征整合，构建了一种层次化的特征表示，这样不同层的输出就可以满足不同的特征需求。Swin Transformer 在包括图像分类、对象检测和语义分割在内的各种视觉任务上实现了最先进的性能。hucao[4]等人也以swin transformer块作为基本单元，提出了swin-unet[4]，将transformer用到了医学图像领域，并取得了不错的效果。我们基于swin-unet提出了一个非常适合人像畸变矫正任务的新网络。

The proposal of transformer~\cite{vaswani2017attention} has been widely used in natural language processing (NLP). Inspired by their outstanding achievements, researchers have gradually applied transformers to the computer vision field recently~\cite{han2020survey,Khan2021TransformersIV}. More impressively, Liu \emph{et al.}~\cite{liu2021swin} proposed an excellent hierarchical transformer structure called Swin Transformer, which is established upon shifted window partitioning mechanism. It has advanced performance on various vision tasks, including image classification, object detection, and semantic segmentation. Hu \emph{et al.}~\cite{r01} also devised a U-shaped transformer block called Swin-Unet, which 
focused on medical image segmentation and achieved surpassing results. Based on these works, we propose a new transformer network that can meet the need for long-distance semantic information of wide-angle portraits correction.

\section{Method}
% For the wide-angle portraits correction network, it is not easy to obtain a large number of labeled samples because of the extremely high cost. In this paper, we devise a novel semi-supervised scheme to boost performance by utilizing both labeled and unlabeled data. As shown in Fig.~\ref{fig:submodels}, the scheme is achieved by two parts: direction and range consistency (DRC) and regression consistency (RC). Moreover, we construct a network called Multi-Scale Swin-Unet (MS-Unet) to further enhance the correction ability. As shown in Fig.~\ref{fig:structure} (a), the MS-Unet mainly contains MSTBs and SFBs. It takes a single distortion image as input and produces horizontal and vertical correction flow maps as intermediate outputs. Then the distortion image is projected into a normal image by the maps. 
 Fig.~\ref{fig:submodels} shows a pipeline of the proposed method. We devise a novel semi-supervised scheme to solve the problem of limited training data by utilizing both labeled and unlabeled data. As shown, we assume a single distortion image as input. Then, we get the correction flow maps and the segmentation mask as intermediate outputs. The correction flow maps are used to project the distortion image into a correction image. The segmentation mask is the bridge between labeled and unlabeled data.  
 
\subsection{Semi-supervised Learning Algorithm}

As shown in Fig.~\ref{fig:submodels}, in our problem settings, we have a set of unlabeled images noted as $U = (I_{th}^{u})$ and a set of labeled images $L = (I_{th}^{l}, F_{th})$, where $F_{th}$ represents the labels. We mix these images and adopt them to train the correction network through the semi-supervised method composed of DRC and RC, which are described in detail below.

\subsubsection{Direction and Range Consistency (DRC)} 

Many existing methods have proved that the estimation accuracy can be further improved by introducing approximate surrogate tasks \cite{r05,meng2021spatial}. Inspired by the success, we attempt to present a surrogate task (segmentation) into the network that is different from the existing fully-supervised wide-angle portraits correction method~\cite{tan2021practical}. In particular, the segmentation mask from the surrogate task can assist the network to construct a novel direction and range consistency learning strategy, which is helpful to improve the accuracy of wide-angle portraits correction. 

This design is mainly motivated by four aspects: 1) The portraits correction flow maps represent the offset and direction of each pixel for correcting distortion images. By introducing the segmentation task to flow maps, the network pays more attention to learning the direction change of each pixel. It is conducive to the network to understand the portraits correction better. 2) If we generate a binary mask, the network will pay more attention to the guiding role of direction but ignore the importance of regional consistency. Thus the multi-category mask is generated by multiple thresholds to supervise the segmentation task. In the segmentation mask, the pixels classified into the same category represent that their values change within the same threshold range. In other words, the segmentation mask is also helpful to guide the network to learn the information of regional consistency, so that the correction flow map predicted by the network will also become smoother. 3) As shown in Fig.~\ref{fig:submodels}, the predicted correction flow map can also be converted to the segmentation mask. Hence the loss function can be constructed between portraits correction and segmentation, making it feasible to introduce unlabeled data for our semi-supervised scheme. 4) Meanwhile, the segmentation mask can be generated without extra costs. It is conducive for the transformation between the flow map and segmentation mask so. In addition, the unlabeled data can be fully utilized when conducting our DRC learning strategy.

Semantic segmentation and portraits correction have similar characteristics, making it possible to learn the consistency between them. When the semantic segmentation is deployed as the surrogate task in this paper, it predicts whether the flow map value $F(i,j)$ meets the given direction and range. We judge the offset by the threshold $\delta \in \mathbb{N}^+$, the pixels whose offset is in the range ($-\infty$, -$\delta$] or [$\delta$, $+\infty$) keep negative or positive directions, and the offset in the range (-$\delta$, $\delta$) are merged into one set which indicates slight movement. The prediction target of the segmentation task is defined as follows:

\begin{equation}
S(i, j)=\left\{\begin{matrix}
0, & if& F(i, j) \leqslant -\delta  \\ 
1, & if& -\delta < F(i, j) < \delta \\ 
2, & if& F(i, j)\geqslant \delta
\end{matrix}\right.,
\end{equation}
where $S(i, j)$ denotes the segmentation mask, $(i, j)$ is the pixel position of mask or flow map, and $\delta$ represents the predefined threshold is set to $5$ in our experiments. 

\begin{figure*}[t]
	\begin{center}
		\includegraphics[width=0.92\linewidth]{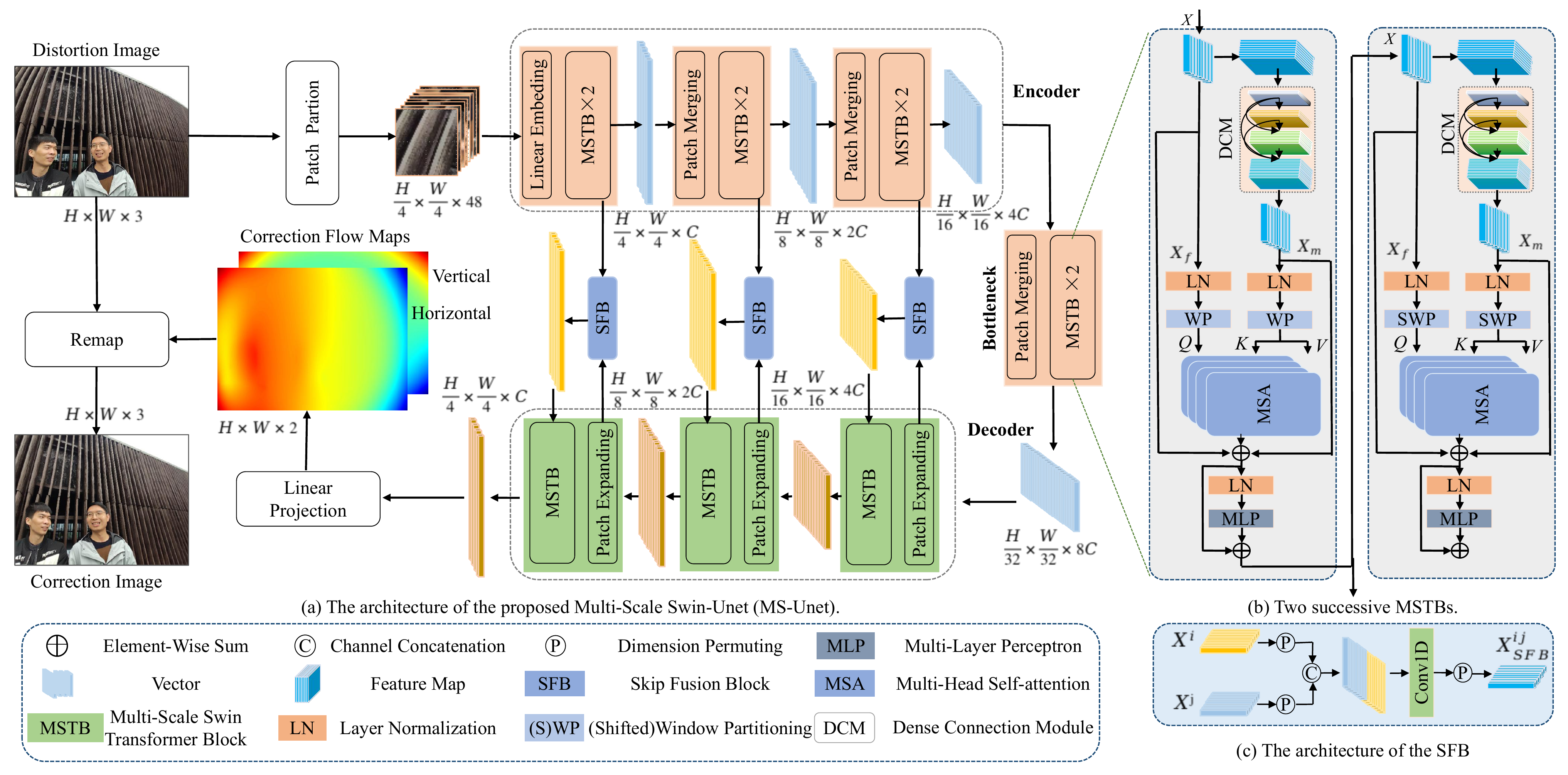}
	\end{center}
	\setlength{\abovecaptionskip}{-0.1cm}
	\caption{(a) The overview of our proposed Multi-Scale Swin-Unet (MS-Unet). The network mainly consists of encoder, decoder, bottleneck and skip fusion blocks (SFB). (b) The architecture of two successive MSTBs. The primary difference between them is the windowing configurations (window partition and shifted window partition). (c) The detailed architecture of SFB. }
	\label{fig:structure}
\end{figure*}

The proposed DRC learning strategy is shown in  Fig.~\ref{fig:submodels}. As mentioned above, by introducing the surrogate task, the network can vigorously supervise direction and regional consistency. For labeled data, the ground truth of the correction flow map is used for training the portraits correction task.  Meanwhile, it also converts into a multi-classification mask, which is utilized for training the surrogate task. As for the unlabeled data, no ground truth is available. Nevertheless, the predicted correction flow map can also generate the segmentation mask through the multiple thresholds. The unlabeled data is still allowed to train the network by DRC loss, and the details will be given in Section \ref{sec:Loss Function}.

\subsubsection{Regression Consistency (RC)} 
Besides the DRC, we also introduce the regression consistency (RC) to improve the network robustness. Fig.\ref{fig:submodels} illustrates the details of RC. Specifically, we can obtain two different images $I_{th}^{u1}$ and $I_{th}^{u2}$ with various augmentation methods (e.g., noise, smoothing, and sharpening), from an unlabeled image $I_{th}^{u}$. Many previous works have stated that an image with different perturbations can obtain similar predictions through a robust network. Therefore, we expand the MS-Unet into a shared-weight siamese structure. The unlabeled images $I_{th}^{u1}$ and $I_{th}^{u2}$ are respectively fed into the two networks, and a consistent loss is established between their outputs. The detailed loss implementation of RC will be given in Section \ref{sec:Loss Function}.

\subsection{Multi-Scale Swin-Unet (MS-Unet)}
\subsubsection{Architecture Overview} 
Although the semi-supervised scheme can significantly boost the performance, the portraits correction depends on a superior network. Motivated by the success of vision transformers~\cite{r01,Khan2021TransformersIV,han2020survey,dosovitskiy2020image}, we develop the MS-Unet, derived from Swin-Unet~\cite{r01}, for the wide-angle portraits correction task. As shown in Fig.~\ref{fig:structure}(a), our proposed MS-Unet can be divided into four major parts: encoder, decoder, bottleneck and skip fusion blocks.

%For the encoder, the input with the size of $H \times W$ is split into non-overlapping patches with the size of $4 \times 4$. Then a linear embedding block is adopted on these patches to form the features with dimension $C$. In addition, there are four hierarchical stages in encoder and bottleneck, and each of them mainly contains two successive MSTBs and a patch merging block. Specially the MSTB is designed for local-scale and long-range information extraction. Contrary to the encoder, the symmetric architecture decoder adopts patch expanding block for up-sampling and only a MSTB in each stage. Before the outputs of each stage are fed into the next stage, they will be fused  with shallow features from encoder via the SFB. Thus the missing spatial information caused by down-sampling will be complemented. Eventually, the final output features from the decoder share the same resolution with the input, and a linear projection layer is employed to produce the horizontal and vertical correction flow maps.

Overall, there are two primary differences between MS-Unet and Swin-Unet. First, as the core unit of Swin-Unet, the swin transformer block ignores the importance of local-scale information, which leads to some objects (e.g., faces with different sizes) being distorted after correction. Second, directly employing the skip connection may not be the optimal scheme for hierarchical features fusion owing to their difference. To alleviate these issues, we leverage the MSTB as the basic unit of our MS-Unet to integrate local-scale and long-range information. Furthermore, the simple yet efficient SFB is designed to replace the skip connection.

\subsubsection{Multi-Scale Swin-Transformer Block (MSTB)} 

Similar to EMSA of RestT\cite{zhang2021rest}, we develop the dense connection module (DCM) into the MSTB for local multi-scale information extraction. In Fig.\ref{fig:structure} (b), two successive MSTBs are presented. Each MSTB contains the DCM, layernorm (LN), multi-head self-attention (MSA), skip connection, and multi-layer perceptron (MLP). The window partitioning (WP) and shifted window partitioning (SWP) are used in two successive MSTBs. 

 When the features $X\in \mathbb{R}^{C\times H\times W}$ with a height of $H$ and a width of $W$ are fed into the MSTB, they will pass through two parallel branches for computing the input of MSA (the query $Q$, key $K$, and value $V$). In the left branch, $X$ are split into non-overlapping windows with a size of $h \times w$ by (S)WP. The features are flattened and reshaped as $X_{f} \in \mathbb{R}^{N\times C}$, where $N = h \times w$. Then a full connection layer is applied to obtain query $Q\in\mathbb{R}^{N\times d}$, where $d = C/k$ and $k$ is the head number. In the right branch, the features $X$ are first utilized to extract local-scale information by DCM. Inspired by  \cite{r03}, the DCM consists of two $1\times1$ layers, and three $3\times3$ depthwise separable convolution layers with different dilation rates $D =(1, 2, 3)$. To be specific, the $1\times1$ convolution layers are employed to change the feature dimension. Each $3\times3$ depthwise separable convolution layer will receive the features from all preceding layers ($i.e,x_{0},...,x_{r-1}$) as input:
\begin{equation}
x_{r} =  C_{r}([x_{0},...,x_{r-1}]),   
\end{equation}
where $C_{r}$ denotes the concatenation operation. Then we apply the same operations like the left branch on these features from the DCM to generate $X_{m} \in \mathbb{R}^{N\times C}$. The key $K\in\mathbb{R}^{N\times d}$, and value $V\in\mathbb{R}^{N\times d}$ are obtained through $X_{m}$.
Afterward, the MSA can be calculated as follows:
\begin{equation}
    MSA(Q, K, V) = Softmax(\frac{QK^{T}}{\sqrt{d}} + B)V,
\end{equation}
where $B \in \mathbb{R}^{N\times N}$ refers to learnable relative position bias.

\subsubsection{Skip Fusion Block (SFB)} 

As mentioned above, the crucial difference among features from different hierarchical stages will be ignored when directly adopting skip connection. Hence, we designed the simple yet efficient skip fusion block (SFB) to replace the skip connection. As shown in Fig. \ref{fig:structure} (c), before the features  $X^{i}\in \mathbb{R}^{N\times C}$, $X^{j}\in \mathbb{R}^{N\times C}$ (from the $i^{th}$ stage of encoder, and the $j^{th}$ stage of decoder) are sent to the next stage of decoder, they pass through the SFB to form new features $X_{SFB}^{ij}$ with dimension  $\mathbb{R}^{N\times C}$. The whole calculation process is defined as follows:
\begin{equation}
X_{SFB}^{ij} = D(CON(C[D(X^{i}), D(X^{j})])),
\end{equation}
where $D(\cdot)$ is dimension permuting, $C[\cdot]$ refers to the concatenation, and $CON(\cdot)$ is 1D convolution layer.

\subsection{Loss Function}
\label{sec:Loss Function}

In practice, the MS-Unet is optimized by adopting the supervised losses on the labeled data $L$, and the semi-supervised losses on the unlabeled data $U$.

\subsubsection{Supervised Loss} Our constructed supervised loss $\mathcal{L}_{s}$ is composed of three parts, including mask-based $L1$ loss $\mathcal{L}_{m1}$, mask-based sobel loss $\mathcal{L}_{ms}$, and the cross-entropy loss $\mathcal{L}_{ce}$. The detailed definitions are described as follows: 

$a)$ $\mathcal{L}_{m1}$ $Loss$: In our method, we introduce the weighted mask, which uses the weight value of the portraits area to be greater than that of the background so that the network will pay more attention to the distorted portraits. Eq. \ref{sec:l1} gives the definition of this loss.

\begin{equation}
 \label{sec:l1}   
 \mathcal{L}_{m1} = \left | F^{'}-F \right| M,
\end{equation}
where $F$ and $F^{'}$ represent the ground truth and estimated flow maps, respectively, $M$ denotes the weighted mask.

$b)$ $\mathcal{L}_{ms}$ $Loss$: In portraits correction, the object edges directly affect the overall visual effects of a correction image. Therefore, we introduce the sobel loss, which can be expressed as follows:
\begin{equation}
\begin{aligned}
    \mathcal{L}_{ms}\!=\!\left[ \left| G_{x}(F^{'})\!- \!G_{x}(F)\right | 
    \!+\!\left |  G_{y}(F^{'})\!-\!G_{y}(F) \right | \right]\! M
\end{aligned},
\end{equation}
where the $G_{x}$ and $G_{y}$ mean the sobel operator in horizontal and vertical direction, respectively.

$c)$ $\mathcal{L}_{ce}$ $Loss$: To supervise the mask generated from the segmentation task, we convert the ground truth flow map into a mask label and deploy the cross-entropy loss. The loss function is defined as follows:
\begin{equation}
    \mathcal{L}_{ce} = Slog(S^{'})+(1-S)log(1-S^{'}),
\end{equation}
where the $S$ is the ground truth mask converted from the flow map, and the $S^{'}$ refer to the estimated mask. To sum up, the training loss for a labeled image is:
\begin{equation}
    \mathcal{L}_{s} = \mathcal{L}_{m1} + \lambda_{1}  \mathcal{L}_{ms} +  \lambda_{2}  \mathcal{L}_{ce},
\end{equation}
where $\lambda_{1}$ and $\lambda_{2}$ are hyper-parameters of $\mathcal{L}_{ms}$ loss and $\mathcal{L}_{ce}$ loss respectively, both being set to 10 in our experiments.

\subsubsection{Semi-Supervised Loss} For an unlabeled image, we construct the unsupervised loss $\mathcal{L}_{u}$ based on the surrogate (segmentation) task and flow map task, which is used to guide the prediction consistency of the network. Specifically, The unsupervised loss contains two parts: the loss of DRC $\mathcal{L}_{DRC}$ and RC $\mathcal{L}_{RC}$.
\begin{equation}
\begin{aligned}
    \mathcal{L}_{u} &=\mathcal{L}_{RC} + \mathcal{L}_{DRC} \\  &=[\mathcal{L}_{m1}(F_{1}^{'},F_{2}^{'}) +  \lambda_{2} \mathcal{L}_{ms}(F_{1}^{'},F_{2}^{'})] + \\ &[\mathcal{L}_{ce}(S_{1}^{'},S_{1}^{''}) +  \mathcal{L}_{ce}(S_{2}^{'},S_{2}^{''})],    
\end{aligned}
\end{equation}
where the $F_{1}^{'}$ and $F_{2}^{'}$ are the estimated flow maps from both the branches of the siamese network, $S_{1}^{'}$ and $S_{2}^{'}$ refer to the segmentation mask converted from $F_{1}^{'}$ and $F_{2}^{'}$, while the $S_{1}^{''}$ and $S_{2}^{''}$ indicate the output of the siamese network.

\begin{figure*}[ht]
	\begin{center}
		\includegraphics[width=0.94\textwidth]{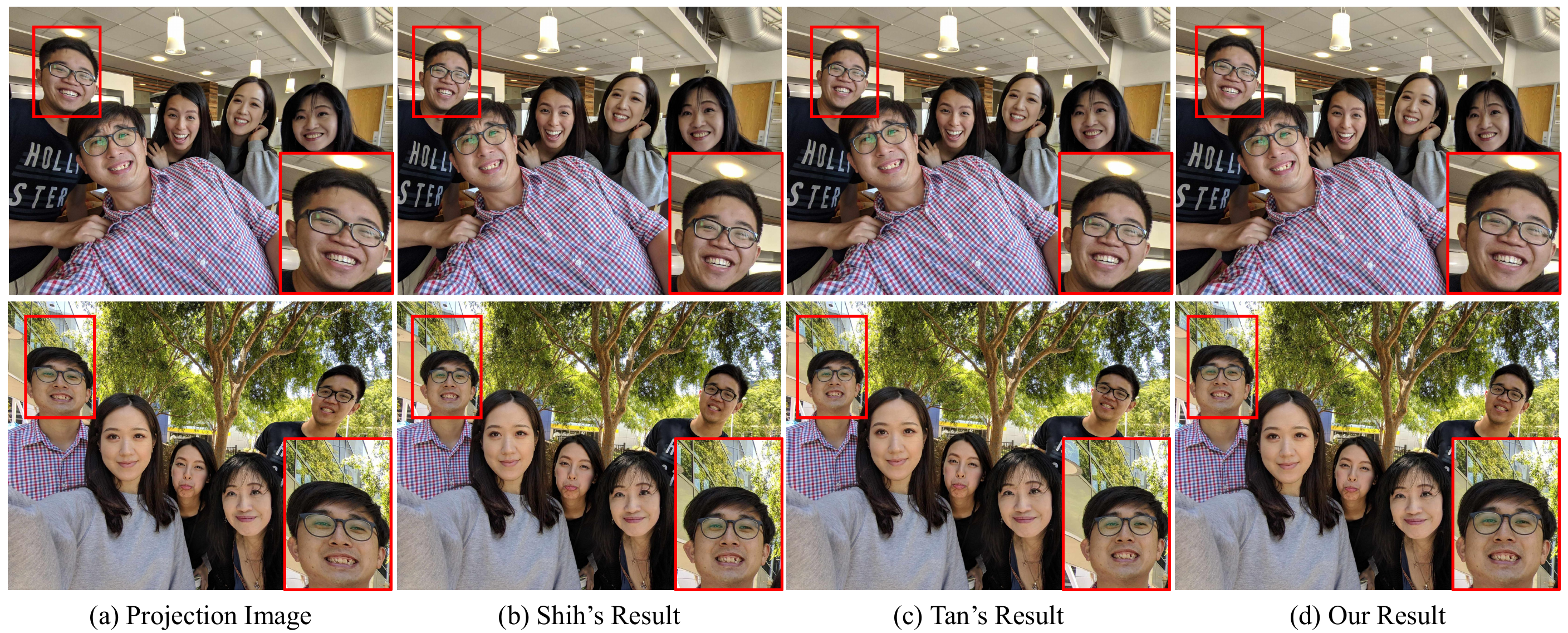}
	\end{center}
	\setlength{\abovecaptionskip}{-0.1cm}
	\caption{Qualitative results of different correction methods. Notice the coordination of lines and face area marked with red boxes.}
	\label{fig:compare_with_other_methods}
\end{figure*}

\section{Experiments}
\subsection{Implementation Details}
\subsubsection{Datasets} 

Following the existing method \cite{tan2021practical}, we conduct extensive experiments on the wide-angle dataset \cite{tan2021practical}, captured with $5$ different smartphones. The training dataset contains over $5,000$ images and $129$ in the testing dataset. Many kinds of labels are provided for each image in the dataset, containing the face mask, correction flow maps, and landmarks. In addition, we collected more than $5, 000$ images by $4$ different smartphones (including Samsung Note 10, Xiaomi 11, vivo X23 and vivo iQOO) as the unlabeled set.

\subsubsection{Training Details} 

%\textcolor{blue}{In the training stage, we train the MS-Unet via a two-step scheduling scheme. Before the semi-supervised strategy starts, we only train correction flow map predictor until $200$ epochs. Then, we introduce the surrogate task and the semi-supervised method to further improve the network's performance further. In this stage, the total training epoch is set to $1,000$ in our experiments. For both steps, we use Adam to optimize our model with a initial learning rate of $1\times10^{-4}$, and the weight decay of $2\times10^{-4}$. In addition, we adopt some data enhancement methods (e.g., zooming, sharpening, smoothing) to enrich the diversity of training samples. All the experiments are performed with $4$ Geforce RTX 2080Ti.}
We train the MS-Unet via a two-step scheme. Similar to~\cite{tan2021practical}, the input size is set as $512\times384$. Before the semi-supervised strategy starts, we need to train a correction flow map predictor, which can provide the pseudo labels for the surrogate task when both labeled and unlabeled images are utilized. We found that the predictor can achieve good results with only $200$ epochs. Then, we introduce the surrogate task (segmentation) to the network, which can enhance the learning ability of the network. Based on the surrogate task, the semi-supervised method can further improve the network's performance. In this stage, the total training epoch is set to $1,000$. Notably, the supervised loss is utilized for labeled images, while the unsupervised loss is for unlabeled images. To compute the loss conveniently, each batch only contains the labeled images or unlabeled images. For both steps, we use Adam to optimize our model with an initial learning rate of $1\times10^{-4}$, and the weight decay of $2\times10^{-4}$. 
%In addition, we adopt some data enhancement methods (e.g., zooming, sharpening, smoothing) to enrich the diversity of training samples. 
%All the experiments are performed with $4$ Geforce RTX 2080Ti.
We trained the MS-Unet ($8.82$ GFLOPs, Parameters: $8.79$M) using $4$ Geforce RTX 2080Ti, and tested it with only one GPU, which can run at around $40$ FPS.

\subsubsection{Evaluation Metrics} 

We use the same evaluation metrics (LineAcc and ShapeAcc) as \cite{tan2021practical} to evaluate the performance of our method. More specifically, LineAcc is used to evaluate the curvature variation of the marked lines and defined as follows:
\begin{equation}
\label{eq:lineacc}
 LS = 1-\frac{1}{n}\sum_{i=0}^{n-1}\left ( \frac{y_{d_{i}} - y_{d_{i-1}}}{x_{d_{i}} - x_{d_{i-1}}} -  \frac{y_{g_{0}} - y_{g_{n}}}{x_{g_{0}} - x_{g_{n}}}\right ),
\end{equation}
where $LS$ denotes the similarity between slope of these two lines, $n$ is the number of uniformly sampled points in each line. $(x_{g_{i}}, y_{g_{i}})$ and $(x_{d_{i}}, y_{d_{i}})$ indicate the coordinate of the corresponding point in the reference and distortion image. 

ShapeAcc aims to evaluate the face similarity between the correction image and the reference image. Based on face landmarks, the ShapeAcc is described as follows:
\begin{equation}
FC =\frac{1}{n}\sum_{i=0}^{n-1}\left \| L_{g_{i}} \right \|\left \| L_{d_{i}} \right \|cos\theta,  
\end{equation}
where $FC$ is the similarity between the corrected and target face, $n$ is the number of fixed sampled points in each face. $L_{g}$ and $L_{d}$ are the corresponding face landmarks in the correction image and the reference image. 

\subsection{Ablation Study} 

In order to verify the influence of different factors on our proposed method, we conducted some ablation experiments on Tan's dataset \cite{tan2021practical} and our unlabeled dataset. Notably, the network structure, the semi-supervised strategy, and the number of unlabeled samples are all considered below.

\subsubsection{Effect of the Correction Network} 

We explore that how the proposed modules affect the network performance using fully-supervised method. Specifically, we utilize the Swin-Unet as our baseline, and the performance of three different networks is evaluated. 1) Baseline: directly employ the Swin-UNet; 2) Baseline+MSTB: based on 1), the MSTB is considered to replace the swin transformer block; 3) Baseline+MSTB+SFB (MS-Unet): the SFB is added to fuse the hierarchical features, and Table \ref{tab1} presents the results. We can observe that the performance boosts significantly with the addition of each module from the table. Oddly, when both MSTB and SFB are added to the network, the full MS-Unet can achieve the best LineAcc ($66.825$) and ShapeAcc ($97.491$). These experiments demonstrate that MSTB indeed promotes the network to extract more complementary information, which boosts the correction ability dramatically. Meanwhile, SFB provides a better feature fusion strategy than skip connections.

In addition, we compared MS-Unet and Tan's method under the same conditions, and the experiment shows that the accuracy of MS-Unet is slightly higher than Tan's network (LineAcc: 66.784, ShapeAcc: 97.490).

\begin{table}[ht]
\centering
\small
\caption{\label{tab1} Ablations on the structure of proposed MS-Unet.}
\begin{tabular}{cccc|cc}
\toprule %添加表格头部粗线
Index & Baseline & MSTB & SFB & LineAcc & ShapeAcc \\
\hline %
1) & \checkmark & - & - & 66.514 & 97.460\\
2) &\checkmark & \checkmark & - & 66.763 & 97.487\\
3) &\checkmark & \checkmark & \checkmark & \textbf{66.825} & \textbf{97.491}\\
\bottomrule
\end{tabular}
\end{table}

\subsubsection{Effect of the Semi-Supervised Strategy}

Several experiments are conducted to evaluate the impact of our proposed semi-supervised scheme. In practice, we first utilize the fully-supervised method to train our MS-Unet with only Tan's dataset~\cite{tan2021practical}. The training result is regarded as the baseline for comparison, and we present it in the first row shown in Table \ref{table: tabsemi}. Then we add the surrogate task to the network and train the two-task MS-Unet. The second row in Table \ref{table: tabsemi} reports the results of the two-task MS-Unet. Compared with the baseline, it shows a slight improvement after adding the surrogate task (LineAcc: $66.825\rightarrow66.871$, ShapeAcc: $97.491\rightarrow97.493$). The result indicates that introducing a surrogate task plays a guiding role in networking training to a certain extent. Afterward, both labeled (Tan's dataset) and unlabeled data are deployed to accomplish the experiments about semi-supervised strategy. The DRC is conducted based on the segmentation task, and the third row in Table \ref{table: tabsemi} lists the comparison result. Compared with the two-task MS-Unet, adding DRC can further improve the estimation accuracy of the correction flow maps, especially the LineAcc (from 66.871 to 67.154). Besides, the effect of RC is also evaluated, and the result is presented in the fourth row of Table \ref{table: tabsemi}. The result also outperforms the single-task MS-UNet, which is only trained by the fully-supervised scheme. The MS-Unet attains the best result (LineAcc: 67.209, ShapeAcc: 97.500) when DRC and RC are employed during the semi-supervised training. These experiment results prove that our proposed semi-supervised strategy can assist the consistency learning between portraits correction task and surrogate and improve the correction performance.

\begin{table}[t]
	\centering
	\small
	\caption{Performance comparison of different semi-supervised strategies. 'Seg' indicates a segmentation task without direction and range consistency. 'DRC' refers to the direction and range consistency, and 'RC' refers to the Regression Consistency.}
	\setlength{\tabcolsep}{6 pt}
% \resizebox{.7\columnwidth}{!}{
	\begin{tabular}{ccccc|cc}
    \toprule %添加表格头部粗线
		  & Baseline & Seg & DRC & RC & LineAcc & ShapAcc\\
		\hline
		 1)  & \checkmark & - & - & - & 66.825 & 97.491\\
		 2)  & \checkmark &\checkmark & - & - & 66.871 & 97.493\\
		 3)  & \checkmark &\checkmark & \checkmark & - & 67.154 & 97.494\\
		 4)  & \checkmark &- & - & \checkmark & 66.848 & 97.497\\
		 5) & \checkmark  & \checkmark & \checkmark & \checkmark & \textbf{67.209} & \textbf{97.500}\\
		\bottomrule
	\end{tabular}
% 	}
	\label{table: tabsemi}
\end{table}

\subsubsection{Effect of the Number of Unlabeled Samples} 

We examine the influence of the number of unlabeled images on network performance through Tan's dataset \cite{tan2021practical} and our unlabeled data. We change the number of unlabeled images from 0 to $5, 000$ while the number of labeled images is fixed. The results are listed in Table \ref{table:numbersemi}, where it shows our MS-Unet trained with semi-supervised strategy obtains consistent superior performance compared with that using the fully-supervised scheme. Meanwhile, we can draw that in a specific range, the performance of the MS-Unet will improve as the amount of unlabeled images increases.

\begin{table}[htbp]
	\centering
	\small
	\caption{The impact of the number of unlabeled images. }
	\begin{tabular}{cc|cc}
   \toprule %添加表格头部粗线
		Index & numbers & LineAcc & ShapeAcc\\
		\hline
		 1) &  0 & 66.871 & 97.493\\
		 2) & 1000 & 66.929 & 97.493\\
		 3) & 2000 & 66.999 & 97.494\\
		 4) & 3000 & 67.105 & 97.497\\
		 5) & 4000 & 67.155 & 97.496\\
		 6) & 5000 & \textbf{67.209} & \textbf{97.500}\\
		\bottomrule
	\end{tabular}
	\label{table:numbersemi}
\end{table}

\begin{figure*}[htbp]
	\begin{center}
		\includegraphics[width=0.95\textwidth]{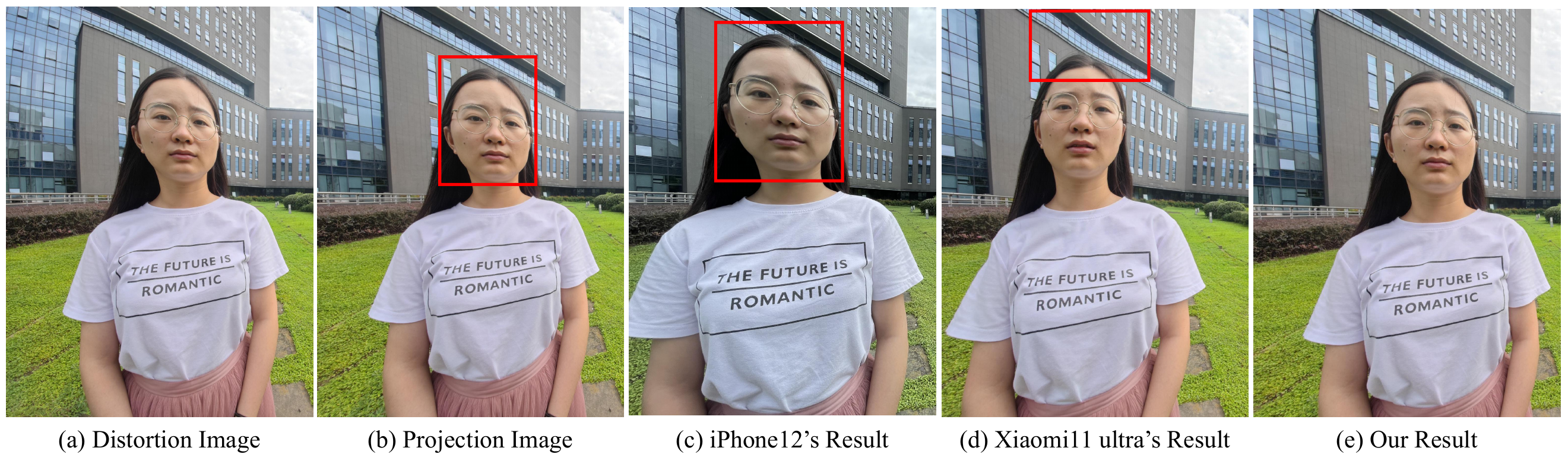}
	\end{center}
	\setlength{\abovecaptionskip}{-0.1cm}
	\caption{Visual comparison between our method and some wide-angle portraits correction methods from smartphones. }
	\label{fig:compare_phones}
\end{figure*}

\subsection{Comparison with Other Methods}

We also compare our method with previous state-of-the-art methods on Tan's and Google's test sets. Table \ref{table:advance} illustrates that our method obtains the highest metric results in both two test sets. The visual comparisons in Fig. \ref{fig:compare_with_other_methods} also confirm the results. Note that the projection image can correct lines but make faces distorted seriously. Shih \cite{shih2019distortion} and Tan \cite{tan2021practical} try to seek the optimal trade-off between the faces and the background. Unfortunately, several bent structures still exist in the background, and a few faces are still distorted. From our results, the faces are more natural than other methods, and the corrected lines in the background are satisfactory. Generally,  both quantitative and qualitative results verify the superior performance of our method.

\begin{table}[t]
    \small
	\centering
	\caption{The results of our proposed method and the two classic methods on two wide-angle portraits correction test sets.}
    \resizebox{.95\columnwidth}{!}{
	\begin{tabular}{c|cc|cc}
    \toprule %添加表格头部粗线
    \multirow{2}{*}{Method} &\multicolumn{2}{c|}{Tan's test set} & \multicolumn{2}{c}{Google's test set} \\
    & LineAcc & ShapeAcc & LineAcc &  ShapeAcc \\
	\hline
	Shih\cite{shih2019distortion} & 66.143 & 97.253 & 61.551 & 97.464 \\
	Tan\cite{tan2021practical} & 66.784 & 97.490 & 64.650 & 97.499 \\
	Ours & \textbf{67.209} & \textbf{97.500} & \textbf{66.098} & \textbf{97.512} \\
	\bottomrule
	\end{tabular}}
	\label{table:advance}
\end{table}

Fig. \ref{fig:compare_phones} depicts the results of our method and other famous portraits correction algorithms from smartphones (i.e., Xiaomi 11 ultra, and iPhone 12). We can observe that serious stretching of portraits appears in iPhone 12. Although Xiaomi 11 ultra improves over the distortion image, there is still slight deformation on the face and curved lines in the background. Our method shows better results, as the face is natural while correcting the lines in the background.

Only a few wide-angle portraits correction works employ the deep learning methods due to its challenge. Based on the correction flow maps, the correction task is regarded as a pixel-level regression problem, which is closely related to some other tasks, such as crowd counting \cite{r05,liu2020semi} and semantic segmentation\cite{chen2018encoder, r01}. Hence, we introduce some efficient networks from these fields to predict the correction flow maps. All the networks are trained by the fully-supervised scheme, and Table \ref{table:computer_vision_semi} shows the results. Notably, our proposed MS-Unet surpasses all the methods. The primary reason is that the CNN-based networks focus on learning local-scale information while the transformer-based networks concentrate on long-range information. For wide-angle portraits, the long-range information can ensure the corrected image generally looks more natural, and the face corrected by local information is more authentic. Therefore, combining both advantages, the MS-Unet will capture multi-scale information for more accurate estimation.

\begin{table}[t]
	\centering
	\small
	\caption{The effectiveness evaluation of the proposed semi-supervised scheme on different networks. }
    \resizebox{1.0\columnwidth}{!}{
	\begin{tabular}{c|cc|cc}
    \toprule %添加表格头部粗线
    \multirow{2}{*}{Method} &\multicolumn{2}{c|}{Fully-Supervised} & \multicolumn{2}{c}{Semi-Supervised} \\
    & LineAcc & ShapeAcc & LineAcc &  ShapeAcc \\
	\hline
	RefineNet\cite{lin2017refinenet} & 66.348 & 97.449 & 66.569 & 97.455 \\
	UNet\cite{r02} & 65.246 & 97.473 & 66.534 & 97.475\\
	CSRNet \cite{li2018csrnet} & 65.967 & 97.469 & 66.236 & 97.471 \\
    Deeplab v3+ \cite{chen2018encoder} & 66.200 & 97.482 & 66.565 & 97.487\\ 
    Swin-Unet\cite{r01} & 66.514 & 97.460 & 66.859 & 97.469\\
    HRNet\cite{sun2019deep} & 66.748 & 97.477 & 66.805 & 97.491\\
	Ours & \textbf{66.825} & \textbf{97.491} & \textbf{67.209} & \textbf{97.500} \\
	\bottomrule
	\end{tabular}}
	\label{table:computer_vision_semi}
\end{table}

Finally, our proposed semi-supervised strategy is used to train these networks. Different from the original network architecture, the surrogate task is added to the network during the training process. And all the semi-supervised results are listed in Table \ref{table:computer_vision_semi}. We can observe that the accuracy of these networks assisted by both labeled and unlabeled data is improved compared with the conventional fully-supervised scheme. The experimental results also demonstrate the generalization ability of our semi-supervised method.

\section{Conclusion}
 In this paper, we develop a novel semi-supervised wide-angle portraits correction method using a multi-scale transformer. By combining DRC and RC in our semi-supervised manner, we can solve the limitations of labeled data and fully utilize unlabeled data. In addition, four kinds of smartphones are adopted to collect unlabeled data. Furthermore, we especially propose the MS-Unet, built upon the MSTB, to capture both local-scale and long-range information for improving artifacts around portraits. Extensive experimental results show that our proposed method is much better than the existing advanced methods and can be popularized in the application of wide-angle portraits correction.
 
 \noindent
 {\bf Acknowledgement} This work was supported by the National Natural Science Foundation of China (NSFC) under grants No.61872067 and No.62172032.
 
%%%%%%%%% REFERENCES
{\small
\bibliographystyle{ieee_fullname}
\bibliography{egbib}

\begin{thebibliography}{10}\itemsep=-1pt

\bibitem{babakhin2019semi}
Yauhen Babakhin, Artsiom Sanakoyeu, and Hirotoshi Kitamura.
\newblock Semi-supervised segmentation of salt bodies in seismic images using
  an ensemble of convolutional neural networks.
\newblock In {\em German Conference on Pattern Recognition}, pages 218--231,
  2019.

\bibitem{r01}
Hu Cao, Yueyue Wang, Joy Chen, Dongsheng Jiang, Xiaopeng Zhang, Qi Tian, and
  Manning Wang.
\newblock Swin-unet: Unet-like pure transformer for medical image segmentation.
\newblock {\em arXiv preprint arXiv:2105.05537}, 2021.

\bibitem{carroll2010artistic}
Robert Carroll, Aseem Agarwala, and Maneesh Agrawala.
\newblock Image warps for artistic perspective manipulation.
\newblock {\em {ACM Trans. Graphics}}, 29(4):1--9, 2010.

\bibitem{carroll2009optimizing}
Robert Carroll, Maneesh Agrawala, and Aseem Agarwala.
\newblock Optimizing content-preserving projections for wide-angle images.
\newblock {\em ACM Trans. Graph.}, 28(3):43, 2009.

\bibitem{chen2018encoder}
Liang-Chieh Chen, Yukun Zhu, George Papandreou, Florian Schroff, and Hartwig
  Adam.
\newblock Encoder-decoder with atrous separable convolution for semantic image
  segmentation.
\newblock In {\em {Proc. ECCV}}, pages 801--818, 2018.

\bibitem{cheng2019semi}
Yong Cheng.
\newblock Semi-supervised learning for neural machine translation.
\newblock In {\em Joint training for neural machine translation}, pages 25--40.
  2019.

\bibitem{dosovitskiy2020image}
Alexey Dosovitskiy, Lucas Beyer, Alexander Kolesnikov, Dirk Weissenborn,
  Xiaohua Zhai, Thomas Unterthiner, Mostafa Dehghani, Matthias Minderer, Georg
  Heigold, Sylvain Gelly, et~al.
\newblock An image is worth 16x16 words: Transformers for image recognition at
  scale.
\newblock {\em arXiv preprint arXiv:2010.11929}, 2020.

\bibitem{du2013changing}
Song-Pei Du, Shi-Min Hu, and Ralph~R Martin.
\newblock Changing perspective in stereoscopic images.
\newblock {\em {IEEE Trans. on Visualization and Computer Graphics}},
  19(8):1288--1297, 2013.

\bibitem{fan2021beyond}
Angela Fan, Shruti Bhosale, Holger Schwenk, Zhiyi Ma, Ahmed El-Kishky,
  Siddharth Goyal, Mandeep Baines, Onur Celebi, Guillaume Wenzek, Vishrav
  Chaudhary, et~al.
\newblock Beyond english-centric multilingual machine translation.
\newblock {\em Journal of Machine Learning Research}, 22(107):1--48, 2021.

\bibitem{r05}
Junyu Gao, Qi Wang, and Xuelong Li.
\newblock Pcc net: Perspective crowd counting via spatial convolutional
  network.
\newblock {\em IEEE Trans. on Circuits and Systems for Video Technology},
  30(10):3486--3498, 2019.

\bibitem{han2020survey}
Kai Han, Yunhe Wang, Hanting Chen, Xinghao Chen, Jianyuan Guo, Zhenhua Liu,
  Yehui Tang, An Xiao, Chunjing Xu, Yixing Xu, et~al.
\newblock A survey on visual transformer.
\newblock {\em arXiv preprint arXiv:2012.12556}, 2020.

\bibitem{he2016dual}
Di He, Yingce Xia, Tao Qin, Liwei Wang, Nenghai Yu, Tie-Yan Liu, and Wei-Ying
  Ma.
\newblock Dual learning for machine translation.
\newblock {\em {Proc. NeurIPS}}, 29:820--828, 2016.

\bibitem{he2019bag}
Tong He, Zhi Zhang, Hang Zhang, Zhongyue Zhang, Junyuan Xie, and Mu Li.
\newblock Bag of tricks for image classification with convolutional neural
  networks.
\newblock In {\em {Proc. CVPR}}, pages 558--567, 2019.

\bibitem{r03}
Gao Huang, Zhuang Liu, Laurens Van Der~Maaten, and Kilian~Q Weinberger.
\newblock Densely connected convolutional networks.
\newblock In {\em {Proc. CVPR}}, pages 4700--4708, 2017.

\bibitem{karamanolakis2019leveraging}
Giannis Karamanolakis, Daniel Hsu, and Luis Gravano.
\newblock Leveraging just a few keywords for fine-grained aspect detection
  through weakly supervised co-training.
\newblock {\em arXiv preprint arXiv:1909.00415}, 2019.

\bibitem{Khan2021TransformersIV}
S. Khan, Muzammal Naseer, Munawar Hayat, Syed~Waqas Zamir, F. Khan, and M.
  Shah.
\newblock Transformers in vision: A survey.
\newblock {\em ArXiv}, abs/2101.01169, 2021.

\bibitem{lee2021salnet}
Ju-Hyoung Lee, Sang-Ki Ko, and Yo-Sub Han.
\newblock Salnet: Semi-supervised few-shot text classification with
  attention-based lexicon construction.
\newblock In {\em {Proc. AAAI}}, pages 13189--13197, 2021.

\bibitem{li2019learning}
Xinzhe Li, Qianru Sun, Yaoyao Liu, Qin Zhou, Shibao Zheng, Tat-Seng Chua, and
  Bernt Schiele.
\newblock Learning to self-train for semi-supervised few-shot classification.
\newblock {\em {Proc. NeurIPS}}, 32:10276--10286, 2019.

\bibitem{li2018csrnet}
Yuhong Li and Xiaofan Zhang.
\newblock Csrnet: Dilated convolutional neural networks for understanding the
  highly congested scenes.
\newblock In {\em {Proc. CVPR}}, pages 1091--1100, 2018.

\bibitem{lin2017refinenet}
Guosheng Lin, Anton Milan, Chunhua Shen, and Ian Reid.
\newblock Refinenet: Multi-path refinement networks for high-resolution
  semantic segmentation.
\newblock In {\em {Proc. CVPR}}, pages 1925--1934, 2017.

\bibitem{liu2020semi}
Yan Liu, Lingqiao Liu, Peng Wang, Pingping Zhang, and Yinjie Lei.
\newblock Semi-supervised crowd counting via self-training on surrogate tasks.
\newblock In {\em {Proc. ECCV}}, pages 242--259, 2020.

\bibitem{liu2021swin}
Ze Liu, Yutong Lin, Yue Cao, Han Hu, Yixuan Wei, Zheng Zhang, Stephen Lin, and
  Baining Guo.
\newblock Swin transformer: Hierarchical vision transformer using shifted
  windows.
\newblock {\em arXiv preprint arXiv:2103.14030}, 2021.

\bibitem{meng2021spatial}
Yanda Meng, Hongrun Zhang, Yitian Zhao, Xiaoyun Yang, Xuesheng Qian, Xiaowei
  Huang, and Yalin Zheng.
\newblock Spatial uncertainty-aware semi-supervised crowd counting.
\newblock In {\em {Proc. ICCV}}, pages 15549--15559, 2021.

\bibitem{pavic2006interactive}
Darko Pavi{\'c}, Volker Sch{\"o}nefeld, and Leif Kobbelt.
\newblock Interactive image completion with perspective correction.
\newblock {\em The Visual Computer}, 22(9-11):671--681, 2006.

\bibitem{r02}
Olaf Ronneberger, Philipp Fischer, and Thomas Brox.
\newblock U-net: Convolutional networks for biomedical image segmentation.
\newblock In {\em International Conference on Medical image computing and
  computer-assisted intervention}, pages 234--241, 2015.

\bibitem{shih2019distortion}
YiChang Shih, Wei-Sheng Lai, Chia-Kai Liang, and Chia-Kai Liang.
\newblock Distortion-free wide-angle portraits on camera phones.
\newblock {\em {ACM Trans. Graphics}}, 38(4):1--12, 2019.

\bibitem{sun2019deep}
Ke Sun, Bin Xiao, Dong Liu, and Jingdong Wang.
\newblock Deep high-resolution representation learning for human pose
  estimation.
\newblock In {\em {Proc. CVPR}}, pages 5693--5703, 2019.

\bibitem{svardal2003stereographic}
Benny~Stale Svardal, Kjell~Einar Olsen, and Odd~Ragnar Andersen.
\newblock Stereographic projection system, Apr.~15 2003.
\newblock US Patent 6,547,396.

\bibitem{tan2021practical}
Jing Tan, Shan Zhao, Pengfei Xiong, Jiangyu Liu, Haoqiang Fan, and Shuaicheng
  Liu.
\newblock Practical wide-angle portraits correction with deep structured
  models.
\newblock In {\em {Proc. CVPR}}, pages 3498--3506, 2021.

\bibitem{tang2017scene}
Youbao Tang and Xiangqian Wu.
\newblock Scene text detection and segmentation based on cascaded convolution
  neural networks.
\newblock {\em IEEE transactions on Image Processing}, 26(3):1509--1520, 2017.

\bibitem{tehrani2013undistorting}
Mahdi~Abbaspour Tehrani, Aditi Majumder, and Meenakshisundaram Gopi.
\newblock Undistorting foreground objects in wide angle images.
\newblock In {\em 2013 IEEE International Symposium on Multimedia}, pages
  46--52. IEEE, 2013.

\bibitem{tehrani2016correcting}
Mahdi~Abbaspour Tehrani, Aditi Majumder, and M Gopi.
\newblock Correcting perceived perspective distortions using object specific
  planar transformations.
\newblock In {\em {Proc. ICCP}}, pages 1--10, 2016.

\bibitem{vaswani2017attention}
Ashish Vaswani, Noam Shazeer, Niki Parmar, Jakob Uszkoreit, Llion Jones,
  Aidan~N Gomez, {\L}ukasz Kaiser, and Illia Polosukhin.
\newblock Attention is all you need.
\newblock In {\em {Proc. NeurIPS}}, pages 5998--6008, 2017.

\bibitem{textseg}
Chuan Wang, Shan Zhao, Li Zhu, Kunming Luo, Yanwen Guo, Jue Wang, and
  Shuaicheng Liu.
\newblock Semi-supervised pixel-level scene text segmentation by mutually
  guided network.
\newblock {\em TIP}, 30:8212--8221, 2021.

\bibitem{xiao2018transferable}
Huaxin Xiao, Yunchao Wei, Yu Liu, Maojun Zhang, and Jiashi Feng.
\newblock Transferable semi-supervised semantic segmentation.
\newblock In {\em {Proc. AAAI}}, 2018.

\bibitem{xie2020self}
Qizhe Xie, Minh-Thang Luong, Eduard Hovy, and Quoc~V Le.
\newblock Self-training with noisy student improves imagenet classification.
\newblock In {\em {Proc. CVPR}}, pages 10687--10698, 2020.

\bibitem{yalniz2019billion}
Weidi Xu, Haoze Sun, Chao Deng, and Ying Tan.
\newblock Variational autoencoder for semi-supervised text classification.
\newblock {\em {Proc. AAAI}}, 2017.

\bibitem{zhang2018context}
Hang Zhang, Kristin Dana, Jianping Shi, Zhongyue Zhang, Xiaogang Wang, Ambrish
  Tyagi, and Amit Agrawal.
\newblock Context encoding for semantic segmentation.
\newblock In {\em {Proc. CVPR}}, pages 7151--7160, 2018.

\bibitem{zhang2021rest}
Qinglong Zhang and Yu-Bin Yang.
\newblock Rest: An efficient transformer for visual recognition.
\newblock {\em Advances in Neural Information Processing Systems}, 34, 2021.

\bibitem{zorin1995correction}
Denis Zorin and Alan~H Barr.
\newblock Correction of geometric perceptual distortions in pictures.
\newblock In {\em Proceedings of the 22nd annual conference on Computer
  graphics and interactive techniques}, pages 257--264, 1995.

\end{thebibliography}
}

\end{document}